\documentclass{article}
\usepackage{microtype,graphicx,booktabs}
\usepackage{hyperref}

\usepackage[preprint]{icml2026}
\makeatletter
\renewcommand{\ICML@preprint}{\textit{Preprint. November 1, 2026.}}
\makeatother
\usepackage{amsmath,amssymb,array,tabularx}
\hypersetup{pdftitle={TEMPO: Time Series Understanding via Discrete Tokenization},pdfauthor={Riccardo Maggioni},pdfsubject={Time-series understanding; presented at ICML FMSD 2026},colorlinks=true,linkcolor=black,citecolor=black,urlcolor=black}
\icmltitlerunning{TEMPO: Time Series Understanding via Discrete Tokenization}
\begin{document}
\twocolumn[
\icmltitle{TEMPO: Time Series Understanding via Discrete Tokenization}
\begin{icmlauthorlist}
\icmlauthor{Riccardo Maggioni}{forgis}
\end{icmlauthorlist}
\icmlaffiliation{forgis}{Forgis}
\icmlcorrespondingauthor{Riccardo Maggioni}{riccardo.maggioni@forgis.com}
\icmlkeywords{Time series, language models, discrete tokenization}
\vskip 0.3in
]
\printAffiliationsAndNotice{Paper presented at \href{https://icml-structured-fm-workshop.github.io/}{ICML FMSD}.}
\begin{abstract}
Time-series interpretation often requires both numerical observations and textual context. Integrating these inputs into a pretrained language model raises a representation question: how should a signal enter the model's token sequence? We present TEMPO, which extends an LLM's vocabulary with discrete codes from a context-aware Transformer tokenizer using finite scalar quantization. A frozen tokenizer maps four-step patches to a 625-code vocabulary; signal tokens, scale statistics, and text are then processed by the LLM's existing self-attention layers. Adaptation updates the new token embeddings and DoRA adapters. With a Qwen3-4B backbone and task-specific fine-tuning, TEMPO achieves macro-F1 scores of 89.0\% on activity recognition, 85.0\% on sleep staging, and 86.0\% on bearing-fault diagnosis. Time-series question answering and qualitative bearing-diagnosis examples further illustrate how this interface combines signal representations with textual context. Together, these results demonstrate the utility of discrete vocabulary extension for time-series understanding.
\end{abstract}
\section{Introduction}
Time-series data record how a system evolves, while text describes the conditions under which those observations should be interpreted. A vibration trace, for example, may have different implications during commissioning and during routine operation. An interpretation therefore often needs to combine temporal patterns with operating conditions, maintenance history, and a user's question.

Task-specific classifiers provide useful predictions, but their output interface is usually a fixed label set. Language models offer a complementary interface: users can ask questions and receive descriptions or explanations. The central challenge is to give a pretrained LLM access to the signal itself. Numerical text serialization, continuous encoder embeddings, and learned discrete tokens make different trade-offs in sequence length, representation, and integration with the language model.

TEMPO studies a discrete interface. A separately pretrained tokenizer converts a normalized signal into contextual codes. These codes become additional vocabulary entries, so text and time-series tokens occupy one input sequence and use the same self-attention layers. Per-channel mean and standard deviation accompany the codes as text. This separates waveform shape from scale information while retaining both as model inputs.

Our contribution is the combination of a context-aware FSQ tokenizer, direct vocabulary extension, and a two-stage adaptation recipe for time-series tasks. The tokenizer attends over the full observed window before quantization, and the LLM is adapted through new embeddings and DoRA adapters. We evaluate this design on activity recognition, sleep staging, bearing-fault diagnosis, and time-series QA, and analyze the tokenizer encoder representations on UCR datasets.

\section{Related Work}
\paragraph{Time-series foundation models.}
Chronos quantizes scalar values for token-based forecasting~\cite{chronos}; TimesFM and Moirai develop pretrained forecasting models~\cite{timesfm,moirai}. Mantis learns representations for classification~\cite{mantis}. These systems primarily produce forecasts, predictions, or embeddings. TEMPO focuses on integrating observed signals with natural-language questions and responses.

\paragraph{Signal-to-LLM integration.}
ChatTS, OpenTSLM, TsLLM, and ITFormer connect time-series encoders to language models through continuous representations, including soft prompts or cross-attention~\cite{chatts,opentslm,tsllm,itformer}. Soft prompts can enter the existing self-attention stream without requiring a separate cross-attention module.

Discrete approaches include scalar binning in Chat-TS~\cite{chatts_quinlan} and learned vector-quantized representations in TokenCast~\cite{tokencast}. TempoGPT is a closely related approach that quantizes temporal embeddings and shares an embedding layer between temporal and textual tokens~\cite{tempogpt}. Shared discrete vocabularies also support multimodal models for vision and audio~\cite{chameleon,spiritlm}.

TEMPO's distinguishing design choice is a Transformer encoder with global attention over the input window followed by finite scalar quantization (FSQ)~\cite{fsq}. Global attention makes the representation of each patch depend on its surrounding signal context, while FSQ maps these representations onto a fixed scalar grid. TEMPO combines this contextual tokenization with direct vocabulary extension and staged LLM adaptation.

\paragraph{Reasoning over time series.}
FactoryBench evaluates LLM reasoning over industrial robotic telemetry through questions about system state, interventions, counterfactuals, and decisions~\cite{factorybench}. In its zero-shot evaluation of six frontier LLMs, no model exceeds 50\% on structured levels or 18\% on decision-making. These results highlight substantial difficulties in reasoning over industrial time series and motivate interfaces that integrate signal representations with textual context.

\section{Method}\label{sec:method}
\begin{figure*}[t]
\centering\includegraphics[width=\textwidth]{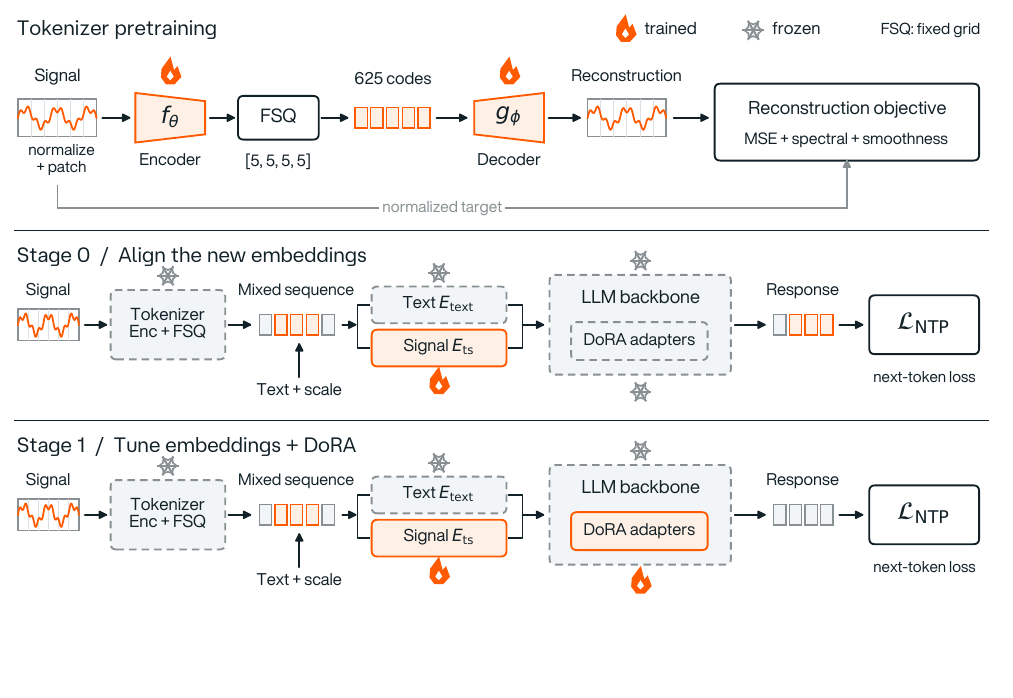}
\caption{\textbf{TEMPO architecture and training stages.} Tokenizer pretraining learns a discrete signal representation through reconstruction. Stage~0 aligns the new signal embeddings with the LLM; Stage~1 adds DoRA adaptation for time-series tasks. Signal codes, scale statistics, and text form a shared input sequence.}\label{fig:architecture}
\end{figure*}

TEMPO comprises a signal tokenizer, an extended LLM vocabulary, and two adaptation stages (Figure~\ref{fig:architecture}). The tokenizer is trained independently of the language model and remains frozen during LLM adaptation.

\paragraph{Preprocessing.}
For channel $c$, we compute its mean $m_c$ and standard deviation $s_c$ over the observed input window and normalize
\begin{equation}\hat{x}_{t,c}=\frac{x_{t,c}-m_c}{\max(s_c,\epsilon)},\end{equation}
where $\epsilon$ prevents division by zero. Each channel is tokenized separately. Its statistics are included in the text context, preserving scale information alongside the normalized shape representation.

\paragraph{Context-aware tokenization.}
The tokenizer divides a normalized channel into non-overlapping patches of $L=4$ time steps and projects them to $d=128$ dimensions. A four-layer Transformer encoder applies global self-attention before a projection to four latent dimensions. FSQ quantizes each dimension to one of five levels, giving an implicit vocabulary of $5^4=625$ codes:
\begin{align}
z_i&=P\!\left(\operatorname{Enc}(\hat{x}_{1:T})_i\right),\\
c_i&=\operatorname{FSQ}(z_i)\in\{0,\ldots,624\}.
\end{align}
Here $i$ indexes patches and $P$ is the latent projection. A symmetric Transformer decoder reconstructs the normalized signal from quantized representations. The tokenizer is pretrained on approximately 500K windows from nine source groups using reconstruction, multi-scale spectral, and temporal-smoothness losses. Training takes approximately four hours on one A10G GPU. Appendix~\ref{app:training} gives the training configuration.

The encoder can represent a patch differently depending on the surrounding signal. This differs from independent scalar binning and from convolutional encoders whose context is limited by their receptive field. The tokenizer compresses four time steps into each code and uses all 625 codes in the vocabulary-usage analysis.

\paragraph{Vocabulary extension.}
We add 627 entries: 625 signal codes and two delimiters, \texttt{<ts\_start>} and \texttt{<ts\_end>}. New embeddings are initialized as random vectors scaled to the mean norm of existing embeddings. A gradient mask restricts embedding updates to the new entries. A typical input has the form
\begin{quote}\small\ttfamily
[user] Signal: (mean=0.12, std=0.45)\\
<ts\_start> <ts\_207> <ts\_14> \ldots\\
<ts\_end> What fault is present?\\
{[assistant]}
\end{quote}
The numbers and token IDs above illustrate the input format. Signal and text tokens use the existing self-attention blocks. The embedding and vocabulary dimensions are expanded, but the Transformer backbone does not require new attention or gating modules.

\paragraph{Stage 0: embedding alignment.}
Only the 627 new embeddings are updated; the tokenizer, backbone, and adapters are frozen. Approximately 200K examples combine discriminative signal tasks (50\%), masked-span code prediction (20\%), and M4 captions (30\%). The discriminative component covers 22 synthetic signal types and 19 task types, with 20\% noise augmentation. This stage associates the discrete signal vocabulary with the language model's existing representation space.

\paragraph{Stage 1: multi-task adaptation.}
We update the new embeddings and DoRA adapters~\cite{dora}, using rank $r=32$ on the attention and MLP projections. The mixture contains approximately 0.26M examples spanning classification, anomaly detection, temporal reasoning, diagnosis, and description. Around 50K text-only instruction examples are included to mitigate language forgetting. Base-model weights and original text embeddings remain frozen.

\paragraph{Task-specific fine-tuning.}
For the in-distribution evaluations, the Stage~1 checkpoint is further fine-tuned on each domain's training split. Each domain is evaluated with its corresponding adapted checkpoint. The tokenizer remains fixed throughout.

\section{Experiments}
\subsection{In-Distribution Evaluation}
We fine-tune the Stage~1 checkpoint on domain-specific training splits and evaluate on held-out test sets. All TEMPO results in Table~\ref{tab:main} use Qwen3-4B. HAR, sleep staging, and bearing diagnosis are evaluated with macro-F1; TSAQA~\cite{tsaqa} classification and anomaly detection use accuracy.

\begin{table*}[t]
\centering\small
\caption{Downstream task results. HAR, Sleep, and Bearing: macro-F1 (\%); TSAQA: accuracy (\%). Models use their respective training protocols. TEMPO results are highlighted in bold.}
\label{tab:main}
\begin{tabular}{llrrrrr}
\toprule
Interface / family & Model & HAR & Sleep & Bearing & TSAQA cls. & TSAQA anom.\\
\midrule
Specialized classifier & Random forest &96.0&63.0&96.1&---&---\\
Specialized classifier & AttnSleep &---&84.4&---&---&---\\
Specialized classifier & WDCNN &---&---&99.2&---&---\\
\midrule
Text serialization (zero-shot) & GPT-4o &2.95$^{\dagger}$&15.5$^{\dagger}$&36.6&28.8&55.7\\
Text serialization (tuned) & LLaMA3.1-8B &---&---&---&91.3&91.0\\
Soft prompt & Llama3.2-3B$^{\dagger}$ &64.87&54.40&---&---&---\\
Cross-attention (Flamingo) & Llama3.2-3B$^{\dagger}$ &62.77&45.45&---&---&---\\
\midrule
Discrete vocabulary (TEMPO) & Qwen3-4B &\textbf{89.0}&\textbf{85.0}&\textbf{86.0}&\textbf{80.5}&\textbf{91.3}\\
\bottomrule
\end{tabular}
\par\smallskip\parbox{\textwidth}{\footnotesize $\dagger$ OpenTSLM macro-F1 values are from Table~2 of \citet{opentslm}, arXiv v3. GPT-4o HAR/Sleep values are also from OpenTSLM; its remaining scores are from local evaluations.}
\end{table*}

TEMPO reaches 89.0\% on HAR, 85.0\% on sleep staging, and 86.0\% on bearing diagnosis. OpenTSLM with Llama3.2-3B is the closest comparison in model size: SoftPrompt reaches 64.87\% on HAR and 54.40\% on Sleep; Flamingo reaches 62.77\% and 45.45\%. TEMPO exceeds both OpenTSLM configurations on these tasks. Specialized models outperform TEMPO on two of the three classification tasks: random forest reaches 96.0\% on HAR, and WDCNN reaches 99.2\% on bearing diagnosis. Closing this gap remains a challenge for the general-purpose signal-to-language interface.

On TSAQA, TEMPO achieves 80.5\% classification accuracy and 91.3\% anomaly-detection accuracy. The tuned LLaMA3.1-8B baseline reaches 91.3\% and 91.0\%, respectively. With a 4B backbone, half the size of the 8B baseline, TEMPO achieves comparable anomaly-detection accuracy but trails it on classification.

\subsection{External-Benchmark Evaluation}
We evaluate transfer to Time-MQA~\cite{timemqa} using a smaller Qwen3-1.7B TEMPO model with one-shot prompting on 600 questions. Table~\ref{tab:external} compares these results with published baselines.

\begin{table}[t]
\centering\small
\caption{Time-MQA accuracy (\%). TEMPO uses one-shot prompting; GPT-4o is zero-shot and Qwen-2.5-7B is a published tuned baseline.}\label{tab:external}
\begin{tabular}{lrrr}
\toprule Model & Anom. & Class. & Judg.\\\midrule
GPT-4o$^\dagger$ &64&32&72\\
Qwen-2.5-7B$^\dagger$ &68&52&82\\
TEMPO (1.7B, one-shot) &56&62&52\\\bottomrule
\end{tabular}
\par\smallskip\footnotesize $\dagger$ Baselines reported in \citet{timemqa}.
\end{table}

TEMPO reaches 62\% classification accuracy, compared with 52\% for Qwen-2.5-7B and 32\% for GPT-4o. Its anomaly-detection and judgment scores are lower than both baselines. The lower judgment score is consistent with the capacity gap between TEMPO's 1.7B backbone and the 7B Qwen baseline. Classification nevertheless remains a strength despite the smaller model size.

\subsection{Cross-Domain Code Sharing}
We analyze code activations on held-out windows from six source groups: ECG, financial, forecasting, synthetic, UCR, and UEA data. The tokenizer uses all 625 codes, with normalized entropy $H/H_{\max}=0.889$. The cosine similarities between domain-level activation vectors are 0.78 for ECG/UCR and 0.73 for financial/forecasting signals.

The activation statistics show overlapping code usage across signal families, with particularly similar distributions for ECG/UCR and financial/forecasting signals.

\subsection{Qualitative: Context-Dependent Responses}
We examine a Qwen3-4B checkpoint adapted to bearing-fault diagnosis using 20K examples. The examples use CWRU signals from an SKF 6205-2RS bearing at 1800 RPM. The following examples illustrate how operational context influences the generated responses.

\paragraph{Context-dependent interpretation.}
For the same vibration trace, the model diagnoses an outer-race fault when told that the bearing is aged. When told that the part is newly installed and non-OEM, it questions whether the observed features reflect a fault or manufacturing variation. The response thus changes with the supplied operating context while the signal remains fixed.

\paragraph{Severity-dependent recommendations.}
Table~\ref{tab:severity} presents four outer-race fault examples. The recommended action escalates with the ISO zone supplied in the prompt, from monitoring to immediate replacement.

\begin{table}[t]
\centering\small
\caption{Bearing-fault examples. Diameter is in inches; RMS is in $g$. ISO zones are supplied textual context, and actions are model-output summaries.}\label{tab:severity}
\begin{tabular}{rrcl}
\toprule Diameter & RMS & Zone & Action summary\\\midrule
0.007 &0.29&A&No immediate action\\
0.014 &0.20&B&Monitor\\
0.021 &0.49&C&Plan replacement\\
0.028 &0.87&D&Replace now\\\bottomrule
\end{tabular}
\end{table}

\paragraph{Response to an incorrect prior.}
In a further example, GPT-4o receives a deliberately incorrect inner-race label together with a numerically serialized signal and produces an explanation consistent with that label. TEMPO instead identifies an outer-race fault in this example.

\section{Discussion and Conclusion}
TEMPO integrates observed time series and text through a shared discrete vocabulary and the LLM's existing self-attention. Its context-aware tokenizer maps four-step patches to a 625-code vocabulary, while textual statistics preserve scale information. This provides a compact signal interface that can be adapted through new embeddings and DoRA adapters while keeping the tokenizer and base LLM weights frozen.

With task-specific adaptation, TEMPO achieves 89.0\%, 85.0\%, and 86.0\% macro-F1 on activity recognition, sleep staging, and bearing diagnosis. The QA results extend this interface to natural-language questions, and the bearing examples illustrate responses conditioned on operating context. The linear-probe analysis in Appendix~\ref{app:probe} shows that the tokenizer encoder representations retain information useful for classification.

\paragraph{Limitations and future work.}
Quantization introduces a trade-off between compression and temporal detail, while separate channel tokenization makes sequence length grow with the number of channels. Future directions include channel-aware compression and adaptation to additional signal domains. FactoryBench~\cite{factorybench} offers a natural extension of the evaluation to industrial questions about interventions, counterfactuals, and decisions.

\bibliographystyle{icml2026}
\bibliography{references}
\clearpage\onecolumn\appendix
\section{Encoder Representation Quality: Linear Probe}\label{app:probe}
We evaluate the tokenizer encoders by mean-pooling their time-series representations and training logistic-regression classifiers on 15 UCR datasets. The encoders remain frozen during probe training. We compare FSQ-T with absolute positional embeddings, CNN+VQ, and FSQ-T with rotary positional embeddings (FSQ-T-RoPE). All 15 probe datasets occur in the tokenizer pretraining corpus, so the comparison measures representation quality under shared dataset exposure. Sample-level overlap is not documented.

\begin{table}[h]
\centering
\caption{Encoder linear-probe accuracy and validation loss.}\label{tab:probe}
\begin{tabular}{lrr}\toprule
Tokenizer & Mean accuracy (\%) & Validation loss\\\midrule
FSQ-T (absolute positions)&\textbf{73.5}&1.42\\
CNN+VQ&71.4&0.94\\
FSQ-T-RoPE&69.6&1.97\\\bottomrule
\end{tabular}
\end{table}

FSQ-T has the highest mean probe accuracy, exceeding CNN+VQ by 2.1 percentage points and FSQ-T-RoPE by 3.9 points. Its validation loss is not the lowest, showing that the ordering by validation loss differs from the ordering by probe accuracy.

\begin{figure}[h]
\centering\includegraphics[width=.94\textwidth]{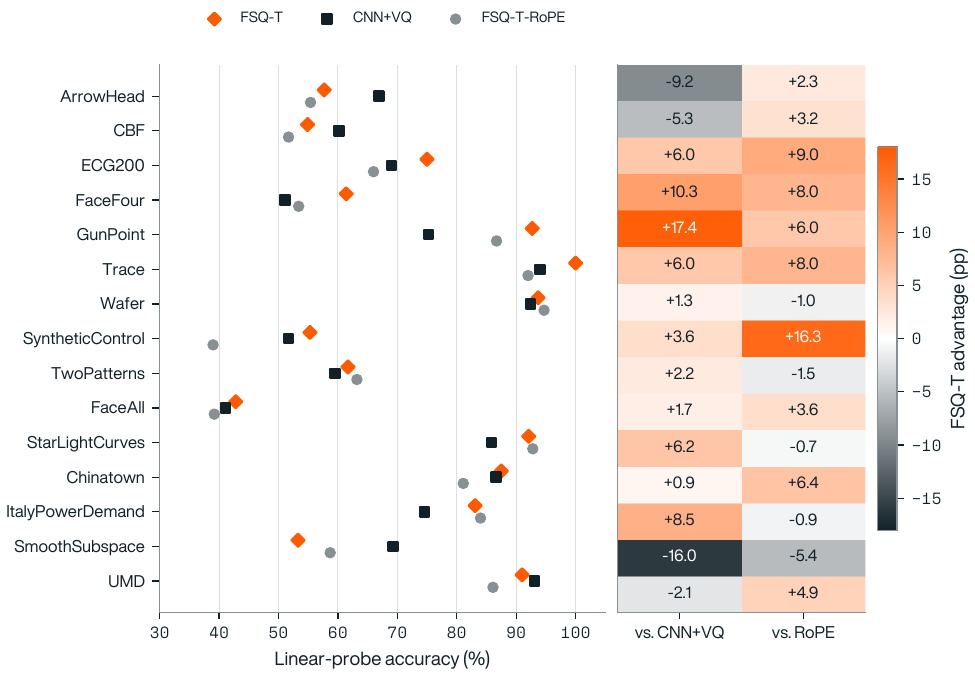}
\caption{Per-dataset probe accuracy and FSQ-T's advantage over each comparator. Values correspond to Table~\ref{tab:probe-detail}. The signed heatmap is centered at zero; orange indicates a positive FSQ-T difference. All 15 datasets were exposed during tokenizer pretraining.}\label{fig:probe}
\end{figure}

\clearpage
\begin{table}[t]
\centering\small
\caption{Per-dataset linear-probe accuracy (\%). Bold marks the highest accuracy in each row; means are unweighted across datasets.}\label{tab:probe-detail}
\begin{tabular}{lrrr}\toprule
Dataset & FSQ-T & CNN+VQ & FSQ-T-RoPE\\\midrule
ArrowHead &57.7&\textbf{66.9}&55.4\\
CBF &54.9&\textbf{60.2}&51.7\\
ECG200 &\textbf{75.0}&69.0&66.0\\
FaceFour &\textbf{61.4}&51.1&53.4\\
GunPoint &\textbf{92.7}&75.3&86.7\\
Trace &\textbf{100.0}&94.0&92.0\\
Wafer &93.7&92.4&\textbf{94.7}\\
SyntheticControl &\textbf{55.3}&51.7&39.0\\
TwoPatterns &61.7&59.5&\textbf{63.2}\\
FaceAll &\textbf{42.8}&41.1&39.2\\
StarLightCurves &92.1&85.9&\textbf{92.8}\\
Chinatown &\textbf{87.5}&86.6&81.1\\
ItalyPowerDemand &83.1&74.6&\textbf{84.0}\\
SmoothSubspace &53.3&\textbf{69.3}&58.7\\
UMD &91.0&\textbf{93.1}&86.1\\\midrule
Mean &\textbf{73.5}&71.4&69.6\\
Dataset wins &7&4&4\\\bottomrule
\end{tabular}
\end{table}

The row-wise maxima give seven wins for FSQ-T, four for CNN+VQ, and four for FSQ-T-RoPE, with no ties at the reported precision. FSQ-T's largest advantage over CNN+VQ is on GunPoint (+17.4 points); its largest disadvantage is on SmoothSubspace (-16.0 points).

\section{Training Details}\label{app:training}
The following tables summarize the training settings. The tokenizer is frozen during both LLM adaptation stages. Stage~0 updates 627 new embeddings; Stage~1 additionally updates DoRA adapters. The expanded vocabulary contains 625 signal codes and two boundary tokens.

\begin{table}[h]
\centering\small
\caption{Model configurations. The 627 new vocabulary entries comprise 625 signal codes and two delimiters.}
\begin{tabular}{lcc}\toprule
Parameter & Qwen3-1.7B & Qwen3-4B\\\midrule
Hidden dimension &2048&3584\\
Transformer layers &28&36\\
Attention heads &16&28\\
New vocabulary entries &627&627\\
DoRA rank / alpha / dropout &32 / 64 / 0.05&32 / 64 / 0.05\\
DoRA target projections &\multicolumn{2}{c}{q, k, v, o, gate, up, down}\\
Signal-code vocabulary &625&625\\
FSQ levels &$[5,5,5,5]$&$[5,5,5,5]$\\
Patch size (time steps) &4&4\\
Maximum signal tokens &750&750\\
Maximum sequence length &2048&2048\\\bottomrule
\end{tabular}
\end{table}

\clearpage
\subsection{Training Phases}
\begin{table}[h]
\centering\small
\caption{Hyperparameters for the main LLM-adaptation runs.}
\begin{tabular}{lll}\toprule
Parameter & Stage 0: alignment & Stage 1: multi-task tuning\\\midrule
Learning rate &$10^{-3}$&$2\times10^{-5}$\\
Epochs &5&3\\
Scheduler &\multicolumn{2}{l}{Linear warmup and linear decay to $0.1\times$ initial rate}\\
Warmup fraction &10\%&10\%\\
Batch size per GPU &8&8\\
Gradient accumulation &2&2\\
Effective batch size &128&128\\
Optimizer &\multicolumn{2}{l}{AdamW, $\beta_1=0.9$, $\beta_2=0.999$}\\
Weight decay &0.01&0.01\\
Precision &BF16 mixed precision&BF16 mixed precision\\
Parallelism &DDP, $8\times$H100 80GB&DDP, $8\times$H100 80GB\\
\bottomrule
\end{tabular}
\end{table}

\paragraph{Stage 0 data.}
The 200K-example mixture contains 50\% discriminative tasks, 20\% code imputation, and 30\% M4 captions. Discriminative tasks use 22 signal types and 19 question types, with 20\% noise augmentation.

\paragraph{Stage 1 data.}
The mixture lists TSAQA-Benchmark (139K), HAR-CoT (20K), Sleep-CoT (9K), CWRU Bearing-CoT (20K), ECG-QA-CoT (20K), and SlimOrca/Alpaca instructions (50K). The listed components total approximately 0.26M examples.

\subsection{Tokenizer Training}
\begin{table}[h]
\centering\small
\caption{Tokenizer configuration and pretraining source categories.}
\begin{tabular}{ll}\toprule
Parameter & Value\\\midrule
Architecture &4-layer Transformer encoder, FSQ, 4-layer Transformer decoder\\
Model dimension / attention heads &128 / 4\\
FSQ grid &$[5,5,5,5]$, 625 codes\\
Patch size / input length &4 / 256 time steps\\
Training corpus &Approximately 500K windows from nine source groups\\
Source categories &Forecasting, UCR, bearing, ECG, financial, IMU, synthetic\\
Objective &MSE + multi-scale spectral + temporal-smoothness losses\\
Optimizer / learning rate &AdamW / $10^{-3}$\\
Hardware / duration &One A10G (24GB) / approximately 4 hours\\
Observed code utilization &625/625 (100\%)\\
Normalized code entropy &$H/H_{\max}=0.889$\\\bottomrule
\end{tabular}
\end{table}

\subsection{Infrastructure and Cost}
Reported costs are approximately \$3 for tokenizer training (four A10G hours), \$49 for Stage~0 (five hours on eight H100s), and \$34 for Stage~1 (3.5 hours on eight H100s). The total is approximately \$86 for the 1.7B pipeline, excluding repeated experiments and downstream specialization.

\subsection{DoRA Rank Observations}
In an exploratory bearing-diagnosis study with Qwen3-4B and 186K training examples, rank 16 produced 9 errors in 30 manually inspected examples. Rank 64 reportedly achieved perfect classification but generated only the bearing template. Rank 32 retained conversational responses with a reported score of 86\%.

\end{document}